\pdfoutput=1

\documentclass[11pt]{article}

\usepackage[]{style/ACL2023}

\usepackage{times}
\usepackage{latexsym}

\usepackage[T1]{fontenc}

\usepackage[utf8]{inputenc}

\usepackage{microtype}
\usepackage{booktabs}
\usepackage{multirow}
\usepackage{amsmath}
\usepackage{amssymb}
\usepackage{graphicx}
\usepackage{balance}
\usepackage{listings}

\usepackage{subcaption}
\usepackage{float}

\usepackage{pifont}

\usepackage[breaklinks]{hyperref}

\usepackage{inconsolata}

\newcommand{\metric}{\textsc{TOWER} }

\title{TOWER: Tree Organized Weighting for Evaluating Complex Instructions}

\author{Noah Ziems, Zhihan Zhang, Meng Jiang \\
  University of Notre Dame \\
  \texttt{\{nziems2, zzhang23, mjiang2\}@nd.edu}
}

\begin{document}
\maketitle
\begin{abstract}

Evaluating the ability of large language models (LLMs) to follow complex human-written instructions is essential for their deployment in real-world applications.
While benchmarks like Chatbot Arena use human judges to assess model performance, they are resource-intensive and time-consuming.
Alternative methods using LLMs as judges, such as AlpacaEval, MT Bench, WildBench, and InFoBench offer improvements but still do not capture that certain complex instruction aspects are more important than others to follow.

To address this gap, we propose a novel evaluation metric, \textsc{TOWER}, that incorporates human-judged importance into the assessment of complex instruction following.
We show that human annotators agree with tree-based representations of these complex instructions nearly as much as they agree with other human annotators.
We release tree-based annotations of the InFoBench dataset and the corresponding evaluation code to facilitate future research.

\end{abstract}

\section{Introduction}
\label{sec:introduction}

\begin{figure}[t]
    \centering
    \includegraphics[width=0.4\textwidth]{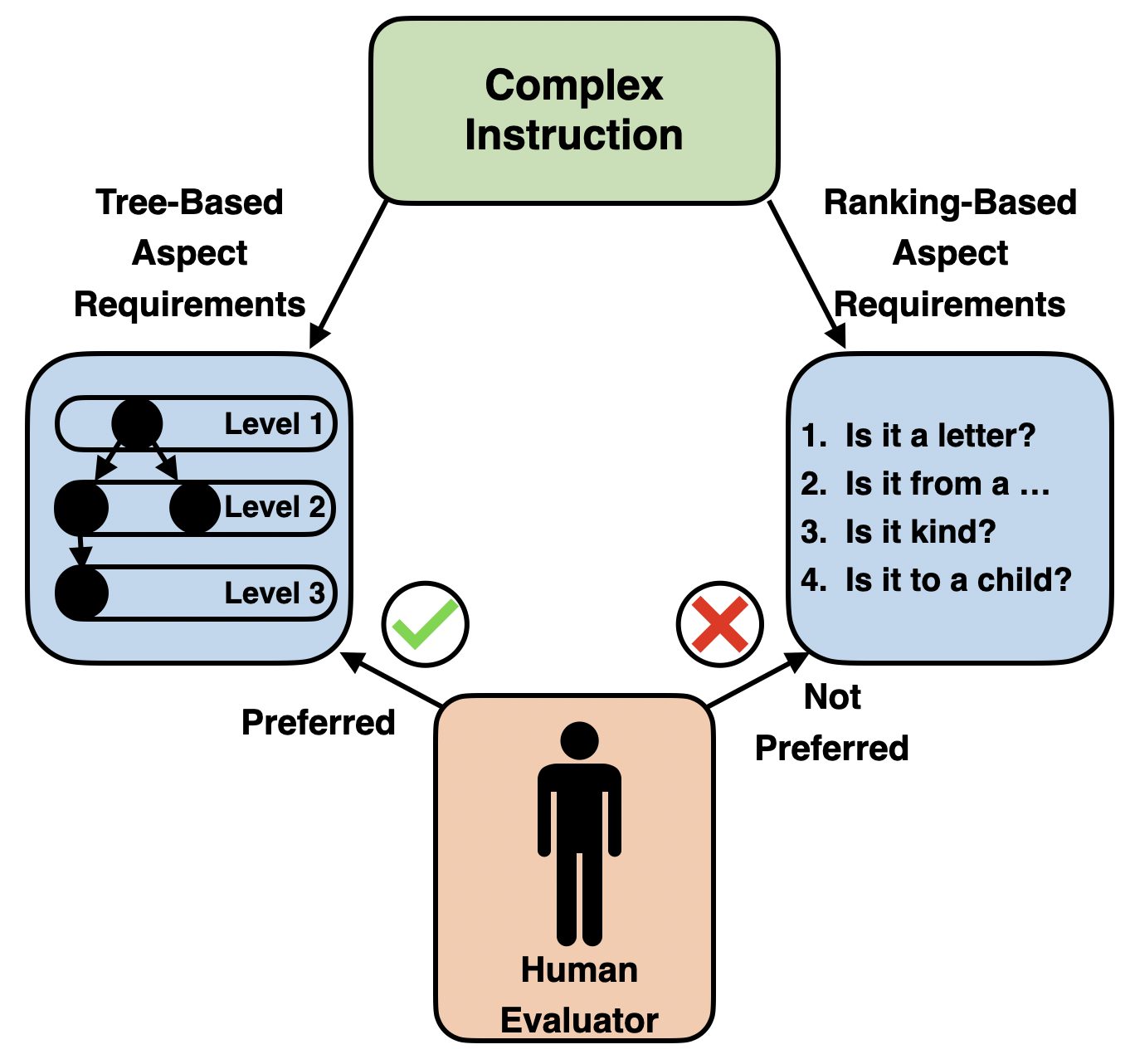}
    \vspace{-0.1in}
    \caption{
    Humans prefer tree-based representations of complex instructions aspect questions over ranking based weighting.
    We distil these insights into a new metric, \textsc{TOWER}, which weights each individual aspect question based on its position within the tree representation.}
    \label{fig:main-figure}
    \vspace{-0.1in}
\end{figure}

The ability of large language models (LLMs) to follow human-written instructions is crucial for real world applications.
Despite this, evaluating and fairly comparing the ability of these LLMs to follow instructions remains a significant challenge, with many proposed approaches.
For instance, Chatbot Arena employs pairwise comparison with crowd sourced human judges from a diverse user base to score the performance of language models \cite{chiang2024chatbot, zheng2023judging}.
Ideally, a model is judged purely on human feedback as in the case of Chatbot Arena, but this is generally expensive, takes lots of time, and is difficult to reproduce.
Further, results can differ based on the selection of the annotators and the distribution of instructions used for evaluation.

Some approaches such as AlpacaEval and MT Bench use LLMs instead of humans as judges, which significantly reduces cost, improves reproducibility, and is much faster \cite{dubois2024length, zheng2023judging, lin2024wildbench}.

However, in practice many real-world use cases involve instructions which are much more complex than the ones found in many instruction following benchmarks.
Some recent promising work has been done creating evaluation benchmarks for these \textit{complex} instructions, which contain many atomic aspects.
One such benchmark, InFoBench, breaks 500 instructions into decomposed requirements, and propose Decomposed Requirements Following Ratio (DRFR) as a metric for measure the percentage of requirements in the benchmark followed by a language model \cite{qin2024infobench}.

Despite these improvements it remains unclear how to best measure performance on complex instruction following, as some aspect of a complex instruction may be more important than others.

In this work, we aim to remedy this issue by proposing a new complex instruction following evaluation metric, \textsc{TOWER}, which weights complex instruction requirements based on human judged importance.


To summarize, our main contributions are as follows:
\begin{enumerate}
    \item We show humans have strong and consistent preferences on the relative importance of different complex instruction aspects.
    \item We show tree-based representations of complex instructions align significantly better with human preferences than ranking or direct scoring, and we use these insights to analyse performance of many existing models.
    \item We release our tree-based annotations of the InFoBench dataset along with the code used to generate them to assist future work in this direction at \url{https://github.com/Ziems/TOWER}
\end{enumerate}
\section{Related Work}
\label{sec:related_work}

The development of language models capable of following human-written instructions has seen significant advancement in recent years.
A large variety of training approaches have been proposed for training these models \cite{touvron2023llama, ouyang2022training, jiang2024mixtral, wang2023far, ivison2023camels, luo2023wizardmath, luo2024wizardcoder, xu2024wizardlm}

Significant research attention has been recently paid to \textit{evaluating} the ability of LLMs to follow instructions.
Some approaches rely on crowd sourcing pairwise human feedback to allow humans to directly judge which model's generation is preferred \cite{chiang2024chatbot}.
Many works have aimed to automate this process by using LLM's instead of humans to judge whether an instruction has been properly followed \cite{chiang2024chatbot, zheng2023judging, dubois2024length}.
Some work has been done to evaluate the ability of language models to follow \textit{complex} instructions which are made of many atomic aspects \cite{qin2024infobench, lin2024wildbench}.

However, relatively little work has been done to evaluate the \textit{importance} of each aspect question within the context of the entire instruction and how these should be measured or evaluated.
\section{Benchmark}
\label{sec:benchmark}


Existing approaches for complex instruction following evaluation such as InFoBench rely on prompting a language model with a complex instruction.
Unlike many prior approaches for evaluation of instruction following, a judge is then asked a series of questions about the generated text, each addressing one aspect of the complex instruction.
For instance, if the instruction asks for a letter from a parent to their child that is warm and supportive, two aspect questions may be \textit{"Is the generated text suitable from a parent to their child?"} and \textit{"Is the letter warm and supportive?"}.
These aspect questions are created by human annotators prior to evaluation and are held constant for each model being evaluated.

Existing metrics for complex instruction following such as Decomposed Requirements Following Ratio (DRFR) from InFoBench weight all instruction aspects equally \cite{qin2024infobench}. 
However, this is not ideal.
In many cases it may be important to weight adherence to higher level aspects such as system-level instructions stronger than adherence to lower level aspects.
To that end, we propose a new instruction aspect weighting scheme which aims to weight the individual aspects of a complex instruction by their relative importance.

As manually annotating the importance of each aspect is a lengthy and prohibitively costly process, we instead opt to use automated annotation and measure agreement with human evaluators.
This also gives us freedom to explore a variety of different weighting approaches with minimal extra cost and to ensure reproducibility.
We propose three different candidate approaches for automated labeling the relative importance of complex instruction aspects: Direct Scoring, Ranking, and Tree-based Weighting.
In the end we find Tree-based importance weighting, which we refer to as 
\textsc{TOWER}, agrees best with human evaluators.
Details on our human annotation process are found in Section \ref{subsec:aspect-importance}.

\begin{table*}[ht]
    \centering
    \begin{tabular}{l|cccc|c}
    \toprule
    \multirow{1}{*}{Weighting Approach} & Annotator 1 & Annotator 2 & Annotator 3 & Annotator 4 & Avg \\
    \midrule
    Human Annotators & 0.78 & 0.70 & 0.70 & 0.80 & 0.74 \\
    Standard DRFR & -0.06 & 0.01 & -0.02 & 0.00 & -0.02 \\
    Ranking & 0.14 & 0.10 & 0.05 & 0.20 & 0.13 \\
    Direct Scoring & 0.36 & 0.16 & 0.23 & 0.37 & 0.28 \\
    \textbf{Tree-Based (Ours) } & \textbf{0.72} & \textbf{0.61} & \textbf{0.73} & \textbf{0.78} & \textbf{0.72} \\
    
    \bottomrule
    \end{tabular}
    \vspace{-0.1in}
    \caption{Annotator agreement with the automated weighting approaches as well as baseline DRFR and agreement between annotators as measured by Spearman correlation. Human annotators are the gold standard. LLM Tree-Based weighting is significantly more aligned with human preferences, agreeing with the human annotators nearly as much as they agree with eachother.}
    \label{tab:annotator-agreement}
    \vspace{-0.05in}
\end{table*}

\subsection{Direct Scoring}
In Direct Scoring, a LLM is directly prompted to score the importance of each aspect on a scale of 1-5, where 5 is the most important and 1 is the least important.
The aspects are all scored in the same generation to ensure that their importance is relative to one another.
It is important to note that, unlike ranking, multiple aspects can have the same score, reflecting that they are of equal importance.

\subsection{Ranking}
Similar to Direct Scoring, in Ranking a LLM is prompted to rank the importance of each aspect relative to each other.
Unlike Direct Scoring, no two aspects can have the same level of importance.

\subsection{Tree-based Weighting}
In Tree-Based Weighting, a tree is constructed by an LLM based on the original complex instruction and the provided aspect questions.
At the root of the tree are the most important aspects with the child and grandchild nodes often being modifications to the original root aspect.
For instance, the aspect questions \textit{"Is the generated text a letter?"} may be the parent of \textit{"Is the generated letter warm and supportive"}.
In this setting, the weight for an aspect question is derived from the level it occurs within the labeled tree, where a parent has higher importance than the children. Specifically the weight is calculated by: $$\frac{1}{level(v)}$$ where $v$ is the node of the aspect question within the tree.

A diagram of this can be found in Figure \ref{fig:main-figure}.



\section{Experiments}
\label{sec:experiment}

Here we discuss our experiments showing that our proposed Tree-Based weighting metric, which we refer from here on as \textsc{TOWER}, strongly aligns with human preferences and provides insight into the complex instruction following capabilities of various models.

\subsection{Dataset}
We use the instructions and aspect questions from InFoBench, a dataset comprised of 500 instructions and 2250 aspect questions with an average of 4.5 aspect questions per instruction.
InFoBench also proposes an evaluation metric, Decomposed Requirements Following Ratio (DRFR), which measures the percentage of aspects which are correctly addressed by a language model's generations.

\subsection{Models}
Following InFoBench, we use a LLM to judge whether a given aspect of an instruction is followed for a particular generation.
We use GPT-4-Turbo for its strong performance, speed, and relatively low cost compared to GPT-4.

We also use GPT-4-Turbo for our experiments involving automated ranking, direct scoring, and tree-based labeling of aspect questions discussed in Section \ref{subsec:aspect-importance}.

A number of different prompts are used to extract the tree structure of the instruction.
We find JSON decoding with a simple example leads to the best quality trees.
Our full prompt can be found in the project repository on GitHub.

\subsection{Human Annotators}

\subsection{Aspect Importance Experiments}
\label{subsec:aspect-importance}

\begin{table*}[ht]
    \vspace{-0.1in}
    \centering
    \begin{tabular}{l|ccccc}
    \toprule
    \multirow{1}{*}{Model} & InFoBench DRFR & \metric (Ours) & Difference \\
    \midrule
    GPT-4-Turbo & 83.56  & 84.48 & +0.92 \\
    Llama-3-70b-instruct & 82.68 & 79.25 & -3.37 \\
    GPT-4-1106 & 81.69 & 84.03 & +2.34 \\
    Claude 2.1 & 80.04 & 76.44 & -3.60 \\
    GPT-3.5-Turbo & 79.56 & 81.90 & +2.34 \\
    Mixtral 8x7B & 53.38 & 60.24 & +6.86 \\
    WizardLM-2 7B & 38.58  & 40.26 & +1.68  \\
    \bottomrule
    \end{tabular}
    \vspace{-0.1in}
    \caption{Standard InFoBench DRFR compared with our Tree-Based Weighting.
    Both metrics use GPT-4-turbo as a judge.
    Larger differences between DRFR and Tree-Based Weighting indicate a model is much better at addressing root aspects than leaf aspects of a complex instruction.}
    \label{tab:model-performance}
    \vspace{-0.05in}
\end{table*}



\begin{figure*}[ht]

\centering
    \begin{subfigure}[b]{0.32\textwidth}
        \centering
        \includegraphics[width=\textwidth]{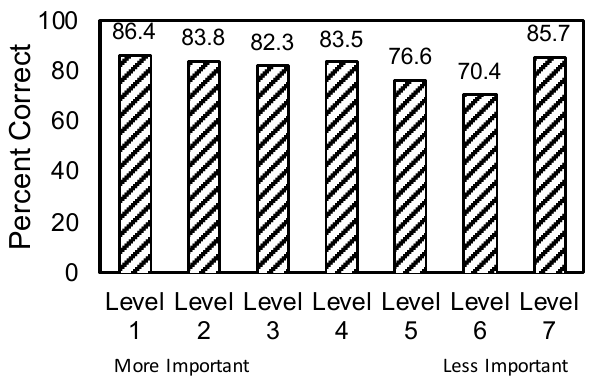}
        \caption{GPT-4-Turbo}
        \label{fig:2a}
    \end{subfigure}
    \hfill
    \begin{subfigure}[b]{0.32\textwidth}
        \centering
        \includegraphics[width=\textwidth]{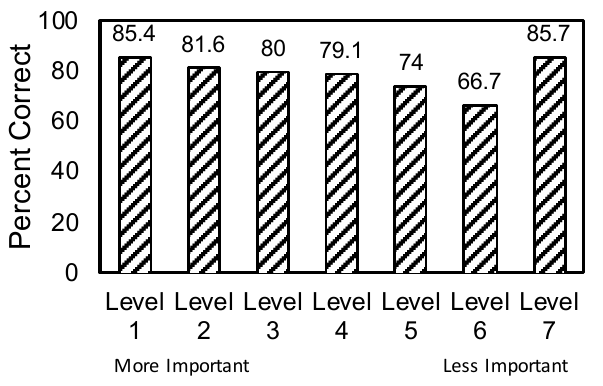}
        \caption{Llama 3-70B}
        \label{fig:2b}
    \end{subfigure}
    \hfill
    \begin{subfigure}[b]{0.32\textwidth}
        \centering
        \includegraphics[width=\textwidth]{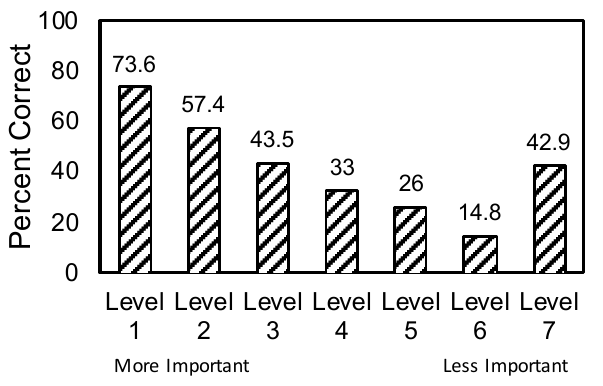}
        \caption{Mixtral 8x7B}
        \label{fig:2c}
    \end{subfigure}
    \caption{Percentage of aspects correctly addressed by model by decision tree level of the individual aspects. GPT-4-Turbo and Llama-3-70B have consistent performance until the 5th level when they begin to degrade. Mixtral performance degrades quickly after the first decision tree level, as indicated in Table \ref{tab:model-performance}}
    \label{fig:2}
     \vspace{-0.1in}
\end{figure*}

To measure the agreement between our LLM weighted aspect importance and human-ranked aspect importance, we use Spearman's rank correlation as follows:
\begin{gather*}
\rho = 1 - \frac{6 \sum d_i^2}{n(n^2 - 1)}
\end{gather*}

where $d_i$ is the difference between the ranks of each pair of instruction aspects and $n$  is the number of instruction aspects.

We use 4 human annotators to independently rank the importance of decomposed questions within the individual instructions.
The annotators are recruited students unfamiliar with the project.
The agreement between annotators and the human weighted aspect importance is shown in Table \ref{tab:annotator-agreement}.

Our results show human annotators have an average Spearman correlation of 0.74 with each other.
Surprisingly, with an average correlation of 0.72, Tree-Based Weighting aligns with human annotators almost as closely as the annotators align with each other.
Ranking and Direct Scoring have substantially lower average correlations with human annotators, at 0.13 and 0.28 respectively.
These results indicate humans annotators strongly prefer Tree-based approaches to instruction aspect importance over ranking or direct scoring.

\subsection{Model Evaluation Results}
We test a number of language models to compare \metric with DRFR.

The model evaluation results show some models are substantially better at addressing all aspects of a complex instruction while others address only the root instruction and struggle to address the lower level aspects.
GPT-4-Turbo has a difference of only 0.92 between the Tree-Based Weighting metric and DRFR, indicating that it addresses all aspects equally and does not tend to over emphasize aspects at the root or the leaves.
Mixtral 8x7B on the other hand has a large difference of 6.86, indicating that it is much better at addressing aspects at the root of the complex instruction tree as opposed than those at the leaves.

To further directly compare DRFR and TOWER, we also experiment by sampling models multiple times with a temperature of 0.8 and use both DRFR and TOWER as the evaluation metric.
We then find the instances where DRFR and TOWER have the largest gap and have human annotators pick which generation has the highest quality.
We find human annotators prefer the same examples as TOWER 68\% of the time, indicating that TOWER aligns much better with human preferences than DRFR.

\subsection{Tree-Based Evaluation Analysis}

Shown in Figure \ref{fig:2}, we further explore the performance gap between DRFR and Tree-Based Weighting of complex instructions.
We show the percentage of aspects correctly addressed for GPT-4-Turbo, Llama-3-70B, and Mixtral 8x7B at the various levels of the complex instruction tree.
GPT-4-Turbo shows a relatively consistent performance across all levels of the decision tree, with performance degradation beginning around Level 5.
Llama-3-70B shows a similar consistent performance, but with degradation beginning instead around Level 4.
Surprisingly, Mixtral 8x7B performance degrades after only the first level, indicating that it does not reliably follow instruction aspects at the lower levels of the trees compared to the higher level ones.
\section{Conclusion and Future Work}
\label{sec:conclusion}

In this paper we explore a new metric for evaluating large language models' ability to follow complex instructions.
Surprisingly, we find tree-based representations have nearly the same Spearman correlation with human annotators (0.72) that human annotators have with themselves (0.74), indicating that tree-based methods are a strong candidate for representing and evaluating complex instructions.
Based on this observation our new proposed metric, \textsc{TOWER}, provides strong insights into which models are better at addressing the most important aspects of complex instructions.

There are numerous exciting directions for future work.
While progress is being made on designing new metrics for evaluating complex instruction following, more benchmark datasets focused on complex instructions are needed as few currently exist in this domain.
Additionally, finding ways to reduce evaluation costs is crucial, as evaluating a single model in InFoBench currently costs approximately \$15.

\section*{Limitations}
\label{sec:limitation}

Despite solving many problems with existing evaluation metrics for complex instruction following, our approach is not without drawbacks.
First and foremost, it is significantly more expensive than InFoBenches DRFR, as pairwise comparison requires many more tokens in the input than single point evaluation.
Further, it is more expensive than AlpacaEval as each individual aspect question requires it's own pairwise evaluation, meaning each instruction will need many pariwise evaluations instead of just one.
As with all evaluation approaches which rely on using LLMs as judges, our approach is susceptible to bias in the LLM, such as preferring longer generations or preferring it's own generations.
\section*{Acknowledgements}
This work was supported by NSF IIS-2119531, IIS-2137396, IIS-2142827, IIS-2234058, CCF-1901059, and ONR N00014-22-1-2507.

\balance
\bibliography{bibliography}

\begin{thebibliography}{13}
\expandafter\ifx\csname natexlab\endcsname\relax\def\natexlab#1{#1}\fi

\bibitem[{Chiang et~al.(2024)Chiang, Zheng, Sheng, Angelopoulos, Li, Li, Zhang, Zhu, Jordan, Gonzalez et~al.}]{chiang2024chatbot}
Wei-Lin Chiang, Lianmin Zheng, Ying Sheng, Anastasios~Nikolas Angelopoulos, Tianle Li, Dacheng Li, Hao Zhang, Banghua Zhu, Michael Jordan, Joseph~E Gonzalez, et~al. 2024.
\newblock Chatbot arena: An open platform for evaluating llms by human preference.
\newblock \emph{arXiv preprint arXiv:2403.04132}.

\bibitem[{Dubois et~al.(2024)Dubois, Galambosi, Liang, and Hashimoto}]{dubois2024length}
Yann Dubois, Bal{\'a}zs Galambosi, Percy Liang, and Tatsunori~B Hashimoto. 2024.
\newblock Length-controlled alpacaeval: A simple way to debias automatic evaluators.
\newblock \emph{arXiv preprint arXiv:2404.04475}.

\bibitem[{Ivison et~al.(2023)Ivison, Wang, Pyatkin, Lambert, Peters, Dasigi, Jang, Wadden, Smith, Beltagy, and Hajishirzi}]{ivison2023camels}
Hamish Ivison, Yizhong Wang, Valentina Pyatkin, Nathan Lambert, Matthew Peters, Pradeep Dasigi, Joel Jang, David Wadden, Noah~A. Smith, Iz~Beltagy, and Hannaneh Hajishirzi. 2023.
\newblock \href {http://arxiv.org/abs/2311.10702} {Camels in a changing climate: Enhancing lm adaptation with tulu 2}.

\bibitem[{Jiang et~al.(2024)Jiang, Sablayrolles, Roux, Mensch, Savary, Bamford, Chaplot, Casas, Hanna, Bressand et~al.}]{jiang2024mixtral}
Albert~Q Jiang, Alexandre Sablayrolles, Antoine Roux, Arthur Mensch, Blanche Savary, Chris Bamford, Devendra~Singh Chaplot, Diego de~las Casas, Emma~Bou Hanna, Florian Bressand, et~al. 2024.
\newblock Mixtral of experts.
\newblock \emph{arXiv preprint arXiv:2401.04088}.

\bibitem[{Lin et~al.(2024)Lin, Deng, Chandu, Brahman, Ravichander, Pyatkin, Dziri, Bras, and Choi}]{lin2024wildbench}
Bill~Yuchen Lin, Yuntian Deng, Khyathi Chandu, Faeze Brahman, Abhilasha Ravichander, Valentina Pyatkin, Nouha Dziri, Ronan~Le Bras, and Yejin Choi. 2024.
\newblock Wildbench: Benchmarking llms with challenging tasks from real users in the wild.
\newblock \emph{arXiv preprint arXiv:2406.04770}.

\bibitem[{Luo et~al.(2023)Luo, Sun, Xu, Zhao, Lou, Tao, Geng, Lin, Chen, and Zhang}]{luo2023wizardmath}
Haipeng Luo, Qingfeng Sun, Can Xu, Pu~Zhao, Jianguang Lou, Chongyang Tao, Xiubo Geng, Qingwei Lin, Shifeng Chen, and Dongmei Zhang. 2023.
\newblock Wizardmath: Empowering mathematical reasoning for large language models via reinforced evol-instruct.
\newblock \emph{arXiv preprint arXiv:2308.09583}.

\bibitem[{Luo et~al.(2024)Luo, Xu, Zhao, Sun, Geng, Hu, Tao, Ma, Lin, and Jiang}]{luo2024wizardcoder}
Ziyang Luo, Can Xu, Pu~Zhao, Qingfeng Sun, Xiubo Geng, Wenxiang Hu, Chongyang Tao, Jing Ma, Qingwei Lin, and Daxin Jiang. 2024.
\newblock \href {https://openreview.net/forum?id=UnUwSIgK5W} {Wizardcoder: Empowering code large language models with evol-instruct}.
\newblock In \emph{The Twelfth International Conference on Learning Representations}.

\bibitem[{Ouyang et~al.(2022)Ouyang, Wu, Jiang, Almeida, Wainwright, Mishkin, Zhang, Agarwal, Slama, Ray et~al.}]{ouyang2022training}
Long Ouyang, Jeffrey Wu, Xu~Jiang, Diogo Almeida, Carroll Wainwright, Pamela Mishkin, Chong Zhang, Sandhini Agarwal, Katarina Slama, Alex Ray, et~al. 2022.
\newblock Training language models to follow instructions with human feedback.
\newblock \emph{Advances in neural information processing systems}, 35:27730--27744.

\bibitem[{Qin et~al.(2024)Qin, Song, Hu, Yao, Cho, Wang, Wu, Liu, Liu, and Yu}]{qin2024infobench}
Yiwei Qin, Kaiqiang Song, Yebowen Hu, Wenlin Yao, Sangwoo Cho, Xiaoyang Wang, Xuansheng Wu, Fei Liu, Pengfei Liu, and Dong Yu. 2024.
\newblock Infobench: Evaluating instruction following ability in large language models.
\newblock \emph{arXiv preprint arXiv:2401.03601}.

\bibitem[{Touvron et~al.(2023)Touvron, Lavril, Izacard, Martinet, Lachaux, Lacroix, Rozi{\`e}re, Goyal, Hambro, Azhar et~al.}]{touvron2023llama}
Hugo Touvron, Thibaut Lavril, Gautier Izacard, Xavier Martinet, Marie-Anne Lachaux, Timoth{\'e}e Lacroix, Baptiste Rozi{\`e}re, Naman Goyal, Eric Hambro, Faisal Azhar, et~al. 2023.
\newblock Llama: Open and efficient foundation language models.
\newblock \emph{arXiv preprint arXiv:2302.13971}.

\bibitem[{Wang et~al.(2023)Wang, Ivison, Dasigi, Hessel, Khot, Chandu, Wadden, MacMillan, Smith, Beltagy, and Hajishirzi}]{wang2023far}
Yizhong Wang, Hamish Ivison, Pradeep Dasigi, Jack Hessel, Tushar Khot, Khyathi~Raghavi Chandu, David Wadden, Kelsey MacMillan, Noah~A. Smith, Iz~Beltagy, and Hannaneh Hajishirzi. 2023.
\newblock \href {http://arxiv.org/abs/2306.04751} {How far can camels go? exploring the state of instruction tuning on open resources}.

\bibitem[{Xu et~al.(2024)Xu, Sun, Zheng, Geng, Zhao, Feng, Tao, Lin, and Jiang}]{xu2024wizardlm}
Can Xu, Qingfeng Sun, Kai Zheng, Xiubo Geng, Pu~Zhao, Jiazhan Feng, Chongyang Tao, Qingwei Lin, and Daxin Jiang. 2024.
\newblock \href {https://openreview.net/forum?id=CfXh93NDgH} {Wizard{LM}: Empowering large pre-trained language models to follow complex instructions}.
\newblock In \emph{The Twelfth International Conference on Learning Representations}.

\bibitem[{Zheng et~al.(2023)Zheng, Chiang, Sheng, Zhuang, Wu, Zhuang, Lin, Li, Li, and Xing}]{zheng2023judging}
L~Zheng, WL~Chiang, Y~Sheng, S~Zhuang, Z~Wu, Y~Zhuang, Z~Lin, Z~Li, D~Li, and E~Xing. 2023.
\newblock Judging llm-as-a-judge with mt-bench and chatbot arena. arxiv preprint arxiv: 230605685.
\newblock \emph{arXiv preprint arXiv:2306.05685}.

\end{thebibliography}
\bibliographystyle{style/acl_natbib}
\nobalance

\appendix
\section{Appendix}

\subsection{Aspect Evaluation Prompt}
\label{subsec:aspect-evaluation-prompt}
For consistency of LLM evaluation we use the same prompt used in InFoBench \cite{qin2024infobench} where an optional input is provided along with the generated text and a final aspect question used for evaluation.
The prompt template for this is shown in Table \ref{tab:infobench-eval}.

\begin{table*}[htbp]
    \small
    \centering
    \renewcommand{\arraystretch}{1.5}
    \begin{tabular}{p{14cm}}
    \lstset{
        basicstyle=\ttfamily\small,
        breaklines=true,
        numbers=none,
        frame=single,  
        framerule=0.8pt,  
        framesep=5pt,  
        backgroundcolor=\color{white},  
        rulecolor=\color{black}  
    }
    \begin{lstlisting}
Based on the provided Input (if any) and Generated Text, answer the ensuing Questions with either a YES or NO choice. Your selection should be based on your judgment as well as the following rules:

- YES: Select `YES' if the generated text entirely fulfills the condition specified in the question. However, note that even minor inaccuracies exclude the text from receiving a `YES' rating. As an illustration, consider a question that asks, "Does each sentence in the generated text use a second person?" If even one sentence does not use the second person, the answer should NOT be `YES'. To qualify for a `YES' rating, the generated text must be entirely accurate and relevant to the question.

- NO: Opt for `NO' if the generated text fails to meet the question's requirements or provides no information that could be utilized to answer the question. For instance, if the question asks, "Is the second sentence in the generated text a compound sentence?" and the generated text only has one sentence, it offers no relevant information to answer the question. Consequently, the answer should be `NO'.

Input: {input}

Generated Text: {model_generation}

Question: {aspect_question}

    \end{lstlisting}
    \end{tabular}
    \caption{Prompt template from InFoBench \cite{qin2024infobench} used for evaluating whether the given model generation follows a given aspect question.}
    \label{tab:infobench-eval}
\end{table*}

\subsection{Tree Decomposition Prompt}
\label{subsec:tree-extraction-prompt}
For decomposing complex instructions into a tree structure, we test many different prompts and find the prompt shown in Table \ref{tab:tree-deconstruction} with JSON decoding works well.

\begin{table*}[htbp]
    \small
    \centering
    \renewcommand{\arraystretch}{1.5}
    \begin{tabular}{p{14cm}}
    \lstset{
        basicstyle=\ttfamily\small,
        breaklines=true,
        numbers=none,
        frame=single,  
        framerule=0.8pt,  
        framesep=5pt,  
        backgroundcolor=\color{white},  
        rulecolor=\color{black}  
    }
    \begin{lstlisting}
I'll provide you with prompts given to a large language model along with aspect questions about the generated output. Your task is to organize the aspect questions into a dependency tree structure without modifying the questions themselves.

An example of a dependency tree structure for the aspect questions is shown below:
```json
{
    "aspect_question": 1,
    "children": [
        {
            "aspect_question": 0,
            "children": []
        },
        {
            "aspect_question": 3,
            "children": []
        },
        {
            "aspect_question": 2,
            "children": []
        },
        {
            "aspect_question": 4,
            "children": []
        },
    ]
}
```

## Instruction

{
    "instruction": """{instruction}""",
    "aspect_questions": """{aspect_questions}""",
}

## Task

Organize the aspect questions into a dependency tree structure without modifying the questions themselves. The tree should be a JSON object with the following format where `aspect_question` is the index of the question in the `aspect_questions` list:
```json
{
    "aspect_question": 0,
    "children": [
        {
            "aspect_question": 1,
            "children": []
        },
        ...
    ]
}
```

## Your answer:
    \end{lstlisting}
    \end{tabular}
    \caption{Prompt template used for deconstructing an instruction and aspect questions into a tree structure.}
    \label{tab:tree-deconstruction}
\end{table*}

\end{document}